\documentclass[sn-mathphys,iicol]{sn-jnl}
\usepackage{comment}
\usepackage{multirow}
\usepackage{bm}
\usepackage{xcolor}
\usepackage{graphicx}
\usepackage{subcaption}

\newcommand{\RED}[1]{\textcolor{black}{#1}}

\jyear{2021}%

\theoremstyle{thmstyleone}%
\theoremstyle{thmstyletwo}

\theoremstyle{thmstylethree}

\begin{document}

\title[Article Title]{Multimodal Skeleton-Based Action Representation Learning via Decomposition and Composition}


\author[1,2]{\fnm{Hongsong} \sur{Wang}}
\author[3]{\fnm{Heng} \sur{Fei}}
\author[3]{\fnm{Bingxuan} \sur{Dai}}
\author[3,4,5]{\fnm{Jie} \sur{Gui}}

\affil[1]{School of Computer Science and Engineering, Southeast University, Nanjing 210096, China}
\affil[2]{Key Laboratory of New Generation Artificial Intelligence Technology and Its Interdisciplinary Applications (Southeast University), Ministry of Education, China}
\affil[3]{School of Cyber Science and Engineering, Southeast University, Nanjing 210096, China}
\affil[4]{Engineering Research Center of Blockchain Application, Supervision And Management (Southeast University), Ministry of Education, China}
\affil[5]{Purple Mountain Laboratories, Nanjing 210000, China}


\abstract{Multimodal human action understanding is a significant problem in computer vision, with the central challenge being the effective utilization of the complementarity among diverse modalities while maintaining model efficiency. However, most existing methods rely on simple late fusion to enhance performance, which results in substantial computational overhead. Although early fusion with a shared backbone for all modalities is efficient, it struggles to achieve excellent performance. To address the dilemma of balancing efficiency and effectiveness, we introduce a self-supervised multimodal skeleton-based action representation learning framework, named Decomposition and Composition. The Decomposition strategy meticulously decomposes the fused multimodal features into distinct unimodal features, subsequently aligning them with their respective ground truth unimodal counterparts. On the other hand, the Composition strategy integrates multiple unimodal features, leveraging them as self-supervised guidance to enhance the learning of multimodal representations. Extensive experiments on the NTU RGB+D 60, NTU RGB+D 120, and PKU-MMD II datasets demonstrate that the proposed method strikes an excellent balance between computational cost and model performance.}




\maketitle

\section{Introduction}
Skeleton-based action recognition has been widely studied. Compared with traditional image- or video-based action recognition techniques \cite{girdhar2019video, feichtenhofer2020x3d, he2021db}, it offers several advantages. First, skeleton sequences eliminate interference from background, lighting, and appearance, and can protect user privacy. Second, skeleton data are relatively sparse, making them more computationally efficient than high-dimensional image and video data. These characteristics indicate a promising future for skeleton-based action understanding.


Owing to the advantages of using skeletal data, many supervised skeleton-based action recognition methods have emerged, achieving high action recognition accuracy. These methods can be categorized into approaches based on RNNs \cite{du2015hierarchical,du2016representation,wang2017modeling}, CNNs \cite{ke2017new}, GCNs \cite{yan2018spatial, shi2020skeleton}, and Transformers \cite{zhang2021stst, shi2020decoupled,zhu2023motionbert, Wang_2023_CVPR2}. Despite extensive research on supervised action recognition \cite{Zhou_2024_CVPR, Wang_2024_CVPR1, Qu_2024_CVPR, Abdelfattah_2024_CVPR, Zhou_2023_CVPR}, its further development is limited by the difficulty of obtaining large amounts of labeled data. 

Self-supervised skeleton-based action recognition, which can make full use of abundant and easily accessible unlabeled data, has come into focus. Techniques such as generative learning \cite{yan2023skeletonmae, mao2023masked} and contrastive learning \cite{10.1145/3581783.3612449, wu2024scd, ijcai2023p95} are prominent examples in this area. Their research focuses on designing special data augmentations \cite{zhang2023hierarchical} or mining more contrastive pairs \cite{dong2023hierarchical, ijcai2023p95, wu2024scd} to enable the model to extract high-level representations through an instance discrimination task.
However, the aforementioned methods are all based on a single modality. 
A few approaches, such as CrosSCLR \cite{li20213d} and CMD \cite{mao2022cmd}, utilize inter-modal interactions during the pretraining phase. 
To leverage multimodal information in skeleton sequences~\cite{wang2018beyond}, nearly all existing approaches use late fusion in a simple manner to combine unimodal predictions\RED{~\cite{shi2019two,song2022constructing}}. This significantly increases the model's complexity and computational cost, leading to challenges in efficiency.

\begin{figure}[t]
	\centering
	\includegraphics[width=0.8\linewidth]{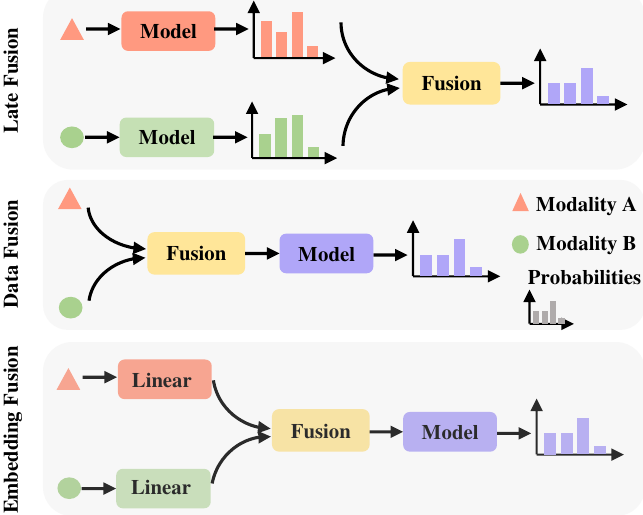}
	\caption{Different multimodal fusion strategies. Late fusion involves the fusion of predicted probabilities or encoded features generated by individual models. Early fusion, on the other hand, entails the fusion directly from raw data, whereas embedding fusion pertains to fusion within the embedded feature space. It should be noted that in this work, early fusion refers to the fusion that occurs at the data level.}
	\label{fig:intro}
\end{figure}

Although early fusion is an efficient way of combining information from different modalities, it often results in a reduction in feature quality. 
Motivated by the training paradigm that learns discriminative multimodal features using a single backbone \cite{10.1145/3581783.3612449}, we present a multimodal skeleton-based action representation learning framework incorporating an embedding fusion strategy. The comparison between late fusion, early fusion, and embedding fusion is shown in Fig.~\ref{fig:intro}. The embedding fusion integrates multimodal information in the embedding space. With carefully designed training methods, the embedding fusion can achieve both efficiency and accuracy for multimodal skeleton-based action understanding.

With this multimodal action representation learning framework, we propose a simple yet effective training method named \textit{Decomposition and Composition}. The \textit{Decomposition} guarantees the inclusion of rich information from various modes within multimodal features by reconstructing the ground-truth unimodal features during training. Since the \textit{Decomposition} approach lacks direct guidance for refining these multimodal features, which may adversely impact the learning of multimodal representations, the \textit{Composition} approach addresses this limitation by directly enhancing the learning of multimodal features with ground-truth multimodal representations.

Specifically, during pretraining, we compose multiple unimodal features and utilize these composed features to augment our multimodal representations. 
This strategy constructs multimodal features by ensemble learning from multiple unimodal features, thereby serving as self-supervised guidance for multimodal representation learning. By leveraging the complementarity among modalities within an embedding fusion framework, the proposed Decomposition and Composition training strategy enables the model to learn both robust unimodal and multimodal action representations.
Additionally, similar to the concept of spatial-temporal decoupling in action recognition \cite{wang2017modeling, zhu2023motionbert}, we incorporate a two-stream feature learning framework. Within this framework, we develop loss functions tailored to both temporal and spatial features, thereby further refining the multimodal representations. We conduct extensive experiments on the NTU RGB+D 60, NTU RGB+D 120, and PKU-MMD II datasets, across various tasks including skeleton-based action recognition, skeleton-based action retrieval, and transfer learning for skeleton-based action recognition. 




In summary, our contributions are as follows:
\begin{itemize}
	\item To the best of our knowledge, our work is one of the few that studies efficient multimodal skeleton-based action representation learning.
	\item We propose a self-supervised multimodal training method called \textit{Decomposition and Composition}, which ensures the effectiveness for both unimodal features and multimodal representations.
	\item With significantly reduced computational demands, our approach achieves state-of-the-art performance on nearly a dozen benchmarks across various tasks, including action recognition and action retrieval.
\end{itemize}

\section{Related Works}


\noindent\textbf{Generative Skeleton-Based Self-Supervised Learning}
Generative self-supervised learning allows the model to learn useful representations by solving tasks such as generating missing parts of an image, predicting the next word in a sentence, or reconstructing noisy data.
In the field of skeleton-based self-supervised learning, there are also researches based on the aforementioned methods.
LongT GAN \cite{zheng2018unsupervised} corrupts the input skeletons and uses an encoder-decoder architecture to reconstruct the original sequences, helping the model understand the structure and dynamics of human motion.
An encoder-decoder recurrent neural network is used to predict and cluster body-keypoints sequences~\cite{su2020predict}.
MS$^2$L \cite{10.1145/3394171.3413548} and HiTRS \cite{chen2022hierarchically} propose multi-task self-supervised learning frameworks, integrating different pretext tasks together like motion prediction and contrastive learning.
Inspired by the success of masked autoencoder \cite{he2022masked} in the image domain, masked skeleton modeling has been introduced into skeleton-based representation learning.
SkeletonMAE \cite{yan2023skeletonmae} and MAMP \cite{mao2023masked} utilize an encoder-decoder structure, where the original input is masked, and the decoder is required to reconstruct the original input from the latent representation generated by the encoder to ensure that it contains information from the original sequence.

\noindent\textbf{Contrastive Skeleton-Based Self-Supervised Learning}
In contrast to reconstruction-based tasks, which focus on fine-grained joint-level details, contrastive learning-based tasks emphasize the global features of skeleton sequences. They learn augmentation-invariant high-level semantic features of actions through instance discrimination tasks, which are independent of data augmentation.
ISC \cite{thoker2021skeleton} utilizes unsupervised contrastive learning to learn representations from different augmentations of skeleton sequences.
Some studies \cite{guo2022contrastive,zhang2023hierarchical} have explored the potential of leveraging stronger data augmentations to benefit contrastive learning of skeleton sequences.
Other works improve representation quality based on motion characteristics, such as local and global relationships \cite{ijcai2023p95,10.1007/978-3-031-19772-7_3,10.1007/978-3-031-19772-7_13} or different relative visual tempo \cite{zhu2023modeling}.
HiCo \cite{dong2023hierarchical} and SCD-Net \cite{wu2024scd} introduce additional features on top of global features and conduct multi-level contrastive learning to enable the model to learn finer-grained positive sample consistencies.
HaLP \cite{Shah_2023_CVPR} reduces the dependence on data augmentation in contrastive learning by constructing positive samples in the latent space for unsupervised learning.
As different modalities of skeleton action sequences can provide complementary semantic information, CrosSCLR \cite{li20213d} and CMD \cite{mao2022cmd} leverage cross-modal information interaction to obtain more comprehensive multimodal information.
ActCLR \cite{lin2023actionlet} divides action sequences into dynamic and static segments in an unsupervised manner to better understand the features.
UmURL \cite{10.1145/3581783.3612449} pretrains the model by decomposing the fused feature into each modality, thus enhancing computational efficiency during the prediction phase.

\noindent\textbf{Multimodal Unsupervised Skeleton-Based Action Recognition} 
In skeleton-based action recognition, skeleton sequences, i.e., joint modal, can be transformed into different modalities, such as motion and bone, to provide diverse human action and body connection information. 
For some previous studies \cite{zhang2023hierarchical, guo2022contrastive, Zhou_Duan_Rao_Su_Wang_2023}, although their focus was not on the interaction between modalities, they utilized and fused outputs from different modalities, such as probabilities or features. Their results showed improvements compared to using unimodal features alone, demonstrating that different modalities can enhance and complement each other.
Meanwhile, some other methods have noted that this complementarity and enhancement can be leveraged during the pretraining phase, where knowledge from different modalities interacts, finally leading to richer and higher-quality multimodal feature representations.
Methods like CrosSCLR \cite{li20213d} and CMD \cite{mao2022cmd} leverage the neighborhood similarity of different modalities as contextual information. 
UmURL \cite{10.1145/3581783.3612449} introduces a unified encoder approach to mitigate the computational overhead associated with three independent encoders. Building on this, we adopt a feature composition method to further enhance the quality of multimodal features, addressing the potential performance loss due to the reduction in the number of parameters.

\section{Decomposition and Composition Training}
\begin{figure*}[t]
	\centering
	\includegraphics[width=0.98\linewidth]{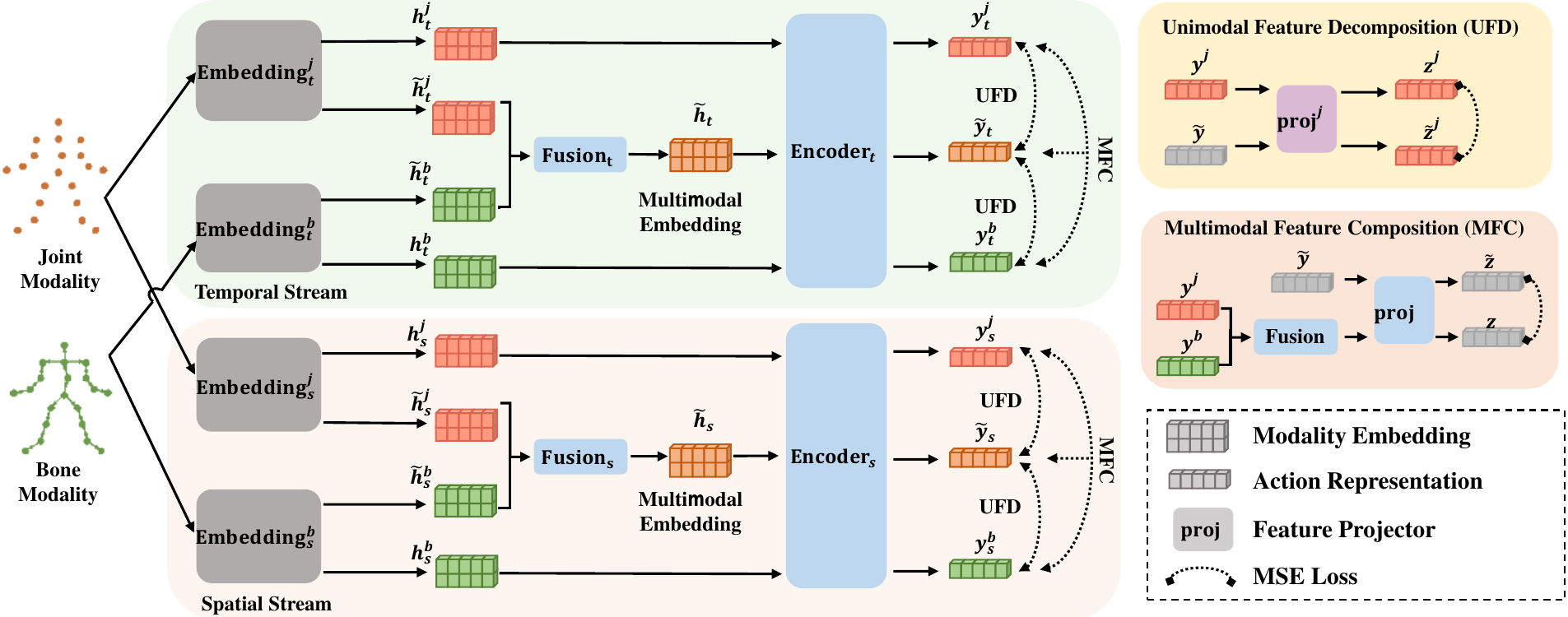}
	\caption{
		The overall pipeline of the proposed method. We use different colors for the skeletons and features to distinguish different input streams of single and unified modalities. In the left part of the figure, it shows that the spatial-temporal encoders of the two branches encode the input separately to obtain decoupled features. The multimodal features are generated using fused embeddings before the encoder. Unimodal Feature Decomposition (UFD) aims to decompose features for more refined comparisons. Multimodal Feature Composition (MFC) aims to build up late fused multimodal features to enhance the features from multimodal embeddings. }
	\label{fig:method}
\end{figure*}

To initiate unified multimodal skeleton-based action representation learning, we first present a concise baseline. Then, we introduce the \text{Decomposition and Composition} training framework in Fig.~\ref{fig:method}, which comprises three modules: Decoupled Spatial-Temporal Encoding, Unimodal Feature Decomposition, and Multimodal Feature Composition.

\subsection{Preliminaries: A Multimodal Baseline}
\noindent\RED{\noindent\textbf{Embedding Fusion}} Different modalities, such as joint, bone, and motion, can be employed to represent a given skeleton sequence. Let $x^k \in\mathbb{R}^{C \times V \times T}$ represent a specific modality, where $k$ is the index of the various modalities, i.e., $k \in \{joint, bone, motion\}$, \(C\) represents the dimension of the coordinates, \(T\) denotes the length of the sequence, \(V\) indicates the number of joints. 

The input from different modalities is mapped to a common high-dimensional embedding space through distinct linear embedding modules:
\begin{equation}
	h^k=\mathrm{{Embedding}}^k(x^k),
\end{equation}
where $h^k\in\mathbb{R}^{D \times V \times T}$, and $D$ indicates the dimension of embedding space. 

To promote multimodal learning and reduce computational complexity, embedding fusion is employed to integrate embeddings from different modalities into a unified embedding. 
\RED{Embeddings from different modalities are fused by either averaging the embeddings directly or applying simple linear transformations to obtain the unified embedding:
\begin{equation}
\label{eq:2}
\tilde{h} =\mathrm{{Fusion}}(h^{joint}, h^{bone}, h^{motion}),
\end{equation}
where $\tilde{h}$ is the unified embedding, $\mathrm{{Fusion}}(\cdot)$ denotes the simple fusion operations, such as averaging.
}

\noindent\RED{\noindent\textbf{Multimodal Action Representations} }
After modality embedding, different embeddings of $h^k$ are fed into the shared skeleton encoder backbone to derive high-level action representations. This unified embedding is also fed into the skeleton encoder to obtain the fused action representation. These steps can be summarized as follows:
\begin{equation}
\label{eq:3}
	y^k=\mathrm{{Encoder}}(h^k), \, \tilde{y} = \mathrm{{Encoder}}(\tilde{h}),
\end{equation}
where $y^k$ is the action representation of the modality $k$, and $\tilde{y}$ denotes the fused action representation of joint, bone, and motion modalities.

It is noteworthy that multimodal features share a common encoder backbone. Since the parameters of the backbone constitute the majority of the total parameters, this approach significantly reduces model parameters in comparison to other multimodal learning approaches that maintain separate backbones for multimodal inputs.

\noindent\RED{\noindent\textbf{Self-Supervised Training} }
Modal-specific projectors are designed to map action representations into a high-dimensional space:
\begin{equation}
	z^k=\mathrm{{proj}}^k(y^k), \, \tilde{z}^k = \mathrm{{proj}}^{k}(\tilde{y}),
\end{equation}
where $\mathrm{{proj}}^{k}(\cdot)$ denotes the feature projector for the modality $k$, $z^k$  represents the mapped representation corresponding to modality $k$, while $\tilde{z}^k$ is the decomposed and estimated representation of modality $k$, derived from the fused representation $\tilde{y}$.

Assuming that the fused action representation can recover the information of the single modality, an unsupervised training loss is constructed by aligning the decomposed feature $\tilde{z}^k$ with the modality feature $z^k$. Using $i$ to index the samples, the Mean Squared Error (MSE) loss is employed to quantify the discrepancy between two features:
\begin{equation}
	\label{dec_g}
	L_{d}=\frac1N\sum_i^N\sum_k^M\|z_{i}^k-\tilde{z}_{i}^{k}\|_2^2.
\end{equation}
where $L_{d}$ is the unsupervised training loss, $M$ and $N$ denote the number of modalities and samples, respectively. \RED{This MSE loss stems from the intra-modal consistency learning in the UmURL~\cite{10.1145/3581783.3612449}.}

In accordance with existing works~
\cite{10.1145/3581783.3612449,weng2025usdrl}, the VC regularization \cite{bardes2022vicreg} is also used to prevent model collapse:
\begin{align}
\label{eq:6}
	V(\bm{Z})=&\frac1{D}\sum_{j=1}^{D}\max(0,\gamma-S(z^j,\epsilon)), \\
	C(\bm{Z})=&\frac{1}{D}\sum_{i}^D\sum_{j \ne i}^D[Cov(\bm{Z})]_{i,j}^{2}.
\label{eq:7}
\end{align}
The regularization $V(\bm{Z})$ ensures that the variance $S(z^j, \epsilon)$ of each column $z^j \in \mathbb{R}^{N}$ in the feature matrix $\bm{Z}\in \mathbb{R} ^{N \times D}$, which represents the value of each feature dimension of a batch, is larger than the hyperparameter $\gamma$, to prevent features from mapping to the same point in space.
The loss function $C(\bm{Z})$ aims to reduce the correlation coefficients between different columns of $\bm{Z}$ to 0, preventing different dimensions from representing the same information and increasing the information in features.

The total training loss can be represented by:
\begin{equation}
\label{eq:8}
	L = L_{d} + \sum_k^M \left (\lambda V(\bm{Z^k}) + \lambda V(\bm{\tilde{Z}^k}) + C(\bm{Z^k}) + C(\bm{\tilde{Z}^k}) \right ),
\end{equation}
where $\lambda$ is a hyperparameter that is usually set to 5, $\bm{Z^k}$ and $\bm{\tilde{Z}^k}$ are feature matrices composed of the outputs $z^k$ and $\tilde{z}^k$ of each modality in a batch.

For the purposes of simplicity and efficiency, this baseline omits the inter-modal consistency loss when compared with UmURL~\cite{10.1145/3581783.3612449}. Our findings indicate that this straightforward alignment of different modalities is ineffective and may even degrade performance.


\subsection{Decoupled Spatial-Temporal Encoding}
Given the skeleton data \(x^k\in\mathbb{R}^{C \times V \times T}\) of the modality $k$, the spatial and temporal views of this data are represented by \(x^{k}_t\in\mathbb{R}^{T \times (V \cdot C)}\) and \(x^{k}_s\in\mathbb{R}^{V \times (T \cdot C)}\), respectively. These data are obtained by retaining the joint and time dimensions respectively, while flattening the other remaining dimensions.

In the temporal branch, the input is processed by an embedding module comprising linear layers and an activation function. This module projects the features of each frame in the sequence into an embedding space. Subsequently, the embedded features are fed into a temporal encoder to obtain the final encoded temporal features $y^{k}_t$. This process is formulated as:
\begin{equation}
	y^{k}_t=\mathrm{{Encoder}_{t}}(\mathrm{{Embedding}}_{t}^{k}(x^{k}_t) )
\end{equation}
where  $\mathrm{{Embedding}}_{t}^{k}(\cdot)$ is temporal embedding layer and $\mathrm{{Encoder}}_t(\cdot)$ is the temporal encoder.

Likewise, spatial features \(y^k_s\) can also be obtained through a similar process, involving modal-specific spatial embedding modules and a spatial transformer encoder.

For unified features, the temporal-specific and spatial-specific features $\tilde{y}_t$ and $\tilde{y}_s$ are also generated according to the method described in Eq. (\ref{eq:2}-\ref{eq:3}).

Same as the baseline model, in the prediction phase, we only use the concatenated unified feature $\tilde{y}=[\tilde{y}_t,\tilde{y}_s]$. However, to obtain more detailed contrast pairs and enhance feature quality, we subsequently use $\tilde{y}^k_t$ and $\tilde{y}_s^k$ for feature decomposing and composition learning.

\subsection{Unimodal Feature Decomposition}
The \textit{Decomposition} training method decomposes the unified features and aligns them with the corresponding modality features separately, ensuring that the multimodal features obtained through embedding fusion explicitly contain information from each modality. 

After obtaining the spatial and temporal representations for each modality and fused modality, the corresponding projection heads are used to map the features. The projection heads are modality-specific. Temporal and spatial branches possess independent projection heads that do not share parameters. Temporal feature decomposition and spatial feature decomposition are designed to independently learn discriminative representations in both temporal and spatial domains. Thus, the loss for unimodal feature decomposition is described as:
\begin{align}
	\label{dec_st}
	L_{d}^t=&\frac1N\sum_i^N\sum_k^M\|z_{i,t}^{k}-\tilde{z}_{i,t}^{k}\|_2^2,\\
	L_{d}^s=&\frac1N\sum_i^N\sum_k^M\|z_{i,s}^{k}-\tilde{z}_{i,s}^{k}\|_2^2.
\end{align}

The overall decomposition loss is expressed as:
\begin{equation}
\label{eq:12}
	L_d=L_d^t+L_d^s.
\end{equation}
By explicitly aligning the temporal and spatial features separately, the approach leverages the inherent spatial-temporal characteristics of skeleton sequences, decomposing features separately to allow for more detailed comparisons. Under this approach, the temporal consistency and spatial consistency between modalities are further ensured.

\subsection{Multimodal Feature Composition}
\RED{To learn effective multimodal action representations, the \textit{Decomposition} loss in Eq.~(\ref{eq:12}) is designed to preserve unimodal characteristics within the fused representations as much as possible. This loss uses multiple unimodal representations as supervisions to guide the learning of multimodal representations.} While the \textit{Decomposition} ensures the incorporation of discriminative unimodal representations, it lacks direct optimization of multimodal features. \RED{A common approach to obtaining multimodal features is through late fusion, where data from different modalities are first independently fed into the backbone network to obtain features, which are subsequently averaged.} In our training framework, multimodal features are obtained by feeding solely the fused embeddings into the backbone. To reduce the discrepancy between these two ways of obtaining multimodal features, we propose a \textit{Composition} training method.



During training, we use late fusion to compose the multimodal features \( z_t \) and \( z_s \) as supervision:
\begin{align}
	\label{mix_st}
	z_t = &\mathrm{{proj}}_{t}\left(\frac{1}{M}\sum_k^M y_t^k\right), \tilde{z}_t =\mathrm{{proj}}_{t}(\tilde{y}_t), \\
	z_s = &\mathrm{{proj}}_{s}\left(\frac{1}{M}\sum_k^M y_s^k\right), \tilde{z}_s =\mathrm{{proj}}_{s}(\tilde{y}_s),
\end{align}
where \( \mathrm{{proj}}_{t}(\cdot) \) and \( \mathrm{{proj}}_{s}(\cdot) \) are specifically used to extract multimodal features, having the same structure as the other projectors. 
$\tilde{z}_t$ and $\tilde{z}_s$ represent the embeddings of fused features in the contrastive space. By bringing these two sets of features closer together in the contrastive space, the composition training loss is formulated as:
\begin{equation}
	L_{c} = \frac{1}{N} \sum_i^N \left(\|z_{i,t} - \tilde{z}_{i,t} \|_2^2 + \|z_{i,s} - \tilde{z}_{i,s} \|_2^2\right).
\end{equation}
We effectively refine the unified modal features for action classification tasks without significantly increasing the computational load during prediction. This approach leverages the strengths of both embedding and late fusion strategies.

\subsection{Training}
\RED{Unlike the VC regularization in the multimodal baseline Eq.~(\ref{eq:6}–\ref{eq:8}), which is only applied to global unimodal features, the improved approach imposes VC regularization on both spatial and temporal unimodal features, as well as on the multimodal features. Thus,} the regularization loss $L_{reg}$ is calculated in a similar manner to the baseline method:
\begin{align}
	L_{reg} & =\sum_k^M \left( L_{vc}(\bm{Z^k_t}) + L_{vc}(\bm{\tilde{Z}^k_t}) + L_{vc}(\bm{Z^k_s}) + L_{vc}(\bm{\tilde{Z}^k_s}) \right) \nonumber \\
	&+ L_{vc}(\bm{Z_t}) + L_{vc}(\bm{Z_s}) + L_{vc}(\bm{\tilde{Z}_t}) + L_{vc}(\bm{\tilde{Z}_s}),
\end{align}
\RED{where $L_{vc}$ is the VC regularization defined in Eq.~(\ref{eq:6}-\ref{eq:7}), $\bm{Z^k_t}$ is the unimodal features after the feature projector for the modality $k$ and the temporal stream, $\bm{Z^k_s}$ is the corresponding feature for the spatial stream, and $\bm{Z^t}$ and $\bm{Z^s}$ are the multimodal features for the temporal and spatial stream, respectively.
}
Each row of $\bm{Z_t^k}$  represents the features $z_t^k$ of each sample in the batch. 
The same applies to the feature matrices $\bm{\tilde{Z}^k_t}$, $\bm{Z_s^k}$, $\bm{\tilde{Z}_s^k}$, $\bm{Z_t}$, $\bm{Z_s}$, $\bm{\tilde{Z}_t}$ and $\bm{\tilde{Z}_s}$.

By integrating the spatial-temporal decomposing and composing item and regularization item, the final loss function can be expressed by:
\begin{equation}
	\label{total}
	L=\alpha L_d+\beta L_{c}+L_{reg}.
\end{equation}
In addition, previous works \cite{dong2023hierarchical,ijcai2023p95,wu2024scd} often treat different data-augmented views of the same sample as positive samples, which enables the model to extract augmentation-invariant features. We refine this training strategy by leveraging ubiquitous multi-viewpoint data. During data collection, obtaining sets of diverse viewpoints for the same action sequence becomes relatively straightforward by simply setting up multiple cameras in order to record human actions simultaneously. 

We devise a viewpoint-invariant training strategy for unified multimodal skeleton-based pretraining. Positive pairs of samples are constructed not only by applying diverse data augmentations to the same sample but also by incorporating the same action sample captured from different viewpoints. Specifically, we represent the positive pairs used for alignment as unordered pairs $(x_i,x_j)$,$1\leq i,j\leq V$, with $V$ denoting the total number of cameras. To utilize viewpoint information, we combine different viewpoints $i$ and $j$ as positive pairs. For samples captured by $V$ cameras, $\frac{V^2+V}{2}$ positive pairs are established, thereby increasing the number of pairs and providing view contrast information.

During training, the input data of the multimodal and unimodal branches come from the same action, which is captured simultaneously by different cameras, but each input undergoes distinct data augmentations. 
This viewpoint information serves as an unsupervised signal that does not require extensive manual annotation and is easy to obtain, almost fully automated, serving as a complement to data augmentation.
By augmenting the samples in this manner, the approach encourages the model to extract useful information that is not only augmentation-invariant but also viewpoint-invariant. 
\section{Experiments}

\subsection{Experimental Settings}
\noindent\textbf{Datasets}
The datasets NTU-60 \cite{shahroudy2016ntu}, NTU-120 \cite{liu2019ntu}, and PKU-MMD II \cite{liu2020benchmark} are employed to validate the proposed method's effectiveness. Specifically, NTU-60 is evaluated using cross-subject (x-sub) and cross-view (x-view) evaluation methods, NTU-120 is evaluated using cross-subject and cross-setup (x-setup) evaluation methods, while PKU-MMD II is evaluated using cross-subject evaluations.

\noindent\textbf{Implementation Details}
The input sequence of the model is truncated to 64 frames and undergoes the same data augmentation procedure as UmURL \cite{10.1145/3581783.3612449}.
In addition to standard data augmentation, the input sequences to the encoder may originate from different or the same samples. The optimization strategy follows the approach adopted in previous work \cite{10.1145/3581783.3612449,mao2022cmd}.
The Adam optimizer \cite{kingma2014adam} with a weight decay of 1e-5 is adopted for optimization. The batch size is 512. The maximum training epochs are 450, 450, and 1000 for the NTU-60, NTU-120, and PKU-MMD datasets, respectively. The learning rate is initially set to 5e-4, and then it is adjusted to 5e-5 at epoch 350 for the NTU-60 and NTU-120 datasets, and at epoch 800 for the PKU-MMD dataset.

\noindent\RED{\textbf{Structure of Action Encoder} Following the two-stream modeling pipeline~\cite{wang2017modeling}, we adopt spatial and temporal transformer networks~\cite{plizzari2021skeleton} as the backbone of the action encoder. Both the spatial and temporal encoders consist of a single Transformer layer with a hidden size of 1024 and one attention head. These encoders are shared across different modalities.
}

\subsection{Comparison with the State-of-the-Arts}
We utilize the model trained in the unsupervised manner mentioned before to obtain action representations and evaluate them using different evaluation protocols across various datasets. The obtained results are then compared with current state-of-the-art methods. For unimodal features, such as joint, we employ the previously mentioned global feature $y^j=[y^j_s,y^j_t]$, while multimodal features are represented by $\tilde{y}=[\tilde{y}_s,\tilde{y}_t]$.

\noindent\textbf{Skeleton-Based Action Recognition}
According to previous work \cite{mao2022cmd,10.1145/3581783.3612449,wu2024scd}, when using the linear evaluation protocol, a single linear layer is added after the backbone model. During training, the parameters of the backbone model are frozen, and only the parameters of the linear layer are updated. 
This method can be used to evaluate the performance of the action features extracted by the model.
Table \ref{tab:linear} presents a comparative analysis of our proposed method against various mainstream approaches on action recognition downstream tasks. Overall, our method demonstrates exceptional performance across different datasets and evaluation metrics. Notably, when integrating joint, motion, and bone modalities, our approach significantly outperforms the existing mainstream methods.
Additionally, the recognition accuracy using only the joint modality is further improved. Compared to UmURL \cite{10.1145/3581783.3612449}, our method achieves greater advancements on the more complex NTU-120 dataset.
In terms of computational cost during the prediction phase, our method remains the same as UmURL, less than other methods, but achieving superior accuracy. Although our model is larger than GCN-based methods, its inference speed on GPU is faster due to the parallelism of the Transformer architecture, especially when processing multimodal skeletons, where it significantly outperforms other approaches in terms of speed.

\begin{table*}[tb]
	\caption{ A comparison of the proposed method with mainstream methods in action recognition.
		`J', `B', and `M' represent joint, bone, and motion modal. The inference time is measured on a single NVIDIA 4090 GPU.
	}
	\renewcommand{\arraystretch}{1.2}
	\centering 
	\scalebox{0.64}{
		\begin{tabular}{lccccccccc}
			\toprule
			\multirow{2}{*}{Method} & \multirow{2}{*}{Modality} & \multirow{2}{*}{\shortstack{FLOPs/(G)}} & \multirow{2}{*}{\shortstack{Model\\Size/M}} & \multirow{2}{*}{FPS} & \multicolumn{2}{c}{NTU-60} & \multicolumn{2}{c}{NTU-120} & \multicolumn{1}{c}{PKU-MMD II} \\
			\cmidrule(lr){6-7} \cmidrule(lr){8-9} \cmidrule(lr){10-10}
			&  &  &  & & x-sub & x-view & x-sub & x-setup & x-sub \\
			\midrule
			AimCLR \cite{guo2022contrastive}  & J & \textbf{1.15} & \textbf{23.2} & 66.6 & 74.3 & 79.7 & 63.4 & 63.4 & - \\
			PSTL \cite{Zhou_Duan_Rao_Su_Wang_2023}  & J & \textbf{1.15} & 23.2 & 66.6 & 77.3 & 81.8 & 66.2 & 67.7 & 49.3 \\
			GL-Transformer \cite{10.1007/978-3-031-19772-7_13}  & J & 118.62 & 1414.2 & -- & 76.3 & 83.8 & 66.0 & 68.7 & - \\
			CPM \cite{10.1007/978-3-031-19772-7_3} & J & 2.22 & -- & -- & 78.7 & 84.9 & 68.7 & 69.6 & 48.3 \\
			CMD \cite{mao2022cmd} & J & 5.76 & 315.5 & 22.2 & 79.8 & 86.9 & 70.3 & 71.5 & 43.0 \\
			ActCLR \cite{lin2023actionlet} &J &1.15& 39.2 & 72.1 & 80.9&86.7&69.0&70.5&-\\
			SkeAttnCLR \cite{ijcai2023p95} & J & 7.32 & 76.2 & \textbf{79.5} & 80.3&86.1&66.3&74.5&\textbf{52.9}\\
			HYSP~\cite{francohyperbolic} &  J & 72.7 & 121.9 & 70.4 & 78.2 & 82.6 & 61.8  & 64.6 & -- \\
			UmURL \cite{10.1145/3581783.3612449} &  J & \underline{1.74} & 277.3 & \underline{78.1} & \underline{82.3} & \underline{89.8} & \underline{73.5} & \underline{74.3} & 52.1 \\
			Ours&J&\underline{1.74}& 277.3 & \underline{78.1} & \textbf{84.1}&\textbf{90.8}&\textbf{75.2}&\textbf{76.7}&\underline{52.7}\\
			\midrule
			3s-AimCLR \cite{guo2022contrastive} & J+M+B & \underline{3.45} & 69.6 & 22.2 & 78.9 & 83.8 & 68.2 & 68.8 & 39.5 \\
			3s-CPM \cite{10.1007/978-3-031-19772-7_3} & J+M+B & 6.66 & -- & -- & 83.2 & 87.0 & 73.0 & 74.0 & 51.5 \\
			3s-CMD \cite{mao2022cmd} & J+M+B & 17.28 & 946.5 & 7.4 & 84.1 & 90.9 & 74.7 & 76.1 & 52.6 \\
			3s-PSTL \cite{Zhou_Duan_Rao_Su_Wang_2023} & J+M+B & \underline{3.45} & 9.66 & 59.2 & 79.1 & 83.8 & 69.2 & 70.3 & 52.3 
			\\
			3s-ActCLR~\cite{lin2023actionlet}&J+M+B& \underline{3.45}& 117.6 & 24.0 &  \underline{84.3}&88.8&74.3&75.7&-\\
			3s-SkeAttnCLR \cite{ijcai2023p95}& J+M+B&21.96& 228.6 & 26.5 & 82.0&86.5&\underline{77.1}&\textbf{80.0}&\textbf{55.5}\\
			3s-UmURL \cite{10.1145/3581783.3612449}& J+M+B & \textbf{2.54} & 277.3 & \textbf{114.1} & 84.2 & \underline{90.9} & 75.2 & 76.3 & 54.0 \\
			Ours & J+M+B &\textbf{2.54}& 277.3 & \textbf{114.1} & \textbf{85.8} & \textbf{91.8} & \textbf{77.5} &\underline{78.8} &\underline{54.7}\\
			\bottomrule
		\end{tabular}
		}
	\label{tab:linear}
\end{table*}

\noindent\textbf{Skeleton-Based Action Retrieval}
The KNN evaluation protocol is another method for assessing the quality of action representations extracted by the model. In the action retrieval task, the model pretrained through contrastive learning is directly used for action retrieval without fine-tuning. The action classification is determined based on the class of the nearest feature.

Table \ref{tab:knn} presents the results of our method on the action retrieval pretext task, comparing it with other methods. Our approach achieves the best performance across all datasets. Notably, on the NTU-60 x-view protocol, our method demonstrates a substantial improvement. The results for the joint modality reach the level of multimodal performance, which we attribute to the view-invariant learning that enables the model to extract invariant semantic information from varying viewpoints.
\begin{table}[tb]
	\caption{A comparison of the proposed method with mainstream methods in action retrieval.
	}
	\renewcommand{\arraystretch}{1.4}
	\centering 
  \scalebox{0.82}{
		\begin{tabular}{@{}l*{6}c @{}}
			\toprule
			\multirow{2}{*}{\textbf{Method}}   & 
			\multirow{2}{*}{\textbf{Modality}}   & 
			\multicolumn{2}{c}{\textbf{NTU-60}} &
			\multicolumn{2}{c}{\textbf{NTU-120}} \\
			\cmidrule(r){3-4} \cmidrule(r){5-6} 
			&&\textbf{x-sub} & \textbf{x-view} & \textbf{x-sub} & \textbf{x-setup} & \\
			\cmidrule{1-6}
			LongT GAN \cite{zheng2018unsupervised} & J & 39.1 & 48.1 & 31.5 & 35.5 & \\
			P\&C \cite{su2020predict} & J & 50.7 & 76.3 & 39.5 & 41.8 & \\
			AimCLR \cite{guo2022contrastive}  & J & 62.0 & 71.5 & - & - & \\
			ISC \cite{thoker2021skeleton} & J & 62.5 & 82.6 & 50.6 & 52.3 & \\
			HiCLR \cite{zhang2023hierarchical} & J & 67.3 & 75.3 & - & - \\
			HiCo \cite{dong2023hierarchical} & J & 68.3 & 84.8 & 56.6 & 59.1\\
			CMD \cite{mao2022cmd} & J & 70.6 & 85.4 & 58.3 &\underline{60.9} \\
			UmURL \cite{10.1145/3581783.3612449} & J & \underline{71.3} & \underline{88.3} & \underline{58.5} &\underline{60.9} \\
			Ours & J &\textbf{72.6} &\textbf{93.0} &\textbf{60.2} &\textbf{63.6}\\
			\hline
			UmURL \cite{10.1145/3581783.3612449}& J+M+B & {72.0} & {88.9} & {59.5} &{62.2} \\
			Ours & J+M+B & \textbf{74.4}&\textbf{93.5} &\textbf{62.4} &\textbf{65.6}\\
			\bottomrule
		\end{tabular}
	}
	\label{tab:knn}
\end{table}

\noindent\textbf{Semi-Supervised Skeleton-Based Action Recognition}
In semi-supervised evaluation, a certain proportion of samples are randomly selected from the training set for finetuning the model, and then action classification tasks are performed on the test set.
This approach allows us to assess the model's performance under conditions with limited labeled data.

To ensure a fair comparison, we fine-tune the UmURL method \cite{10.1145/3581783.3612449} using the same configuration of semi-supervised learning as ours, and the result is shown in Table \ref{tab:semisupervised}.

In single modality J and multi modality J+M+B, our method demonstrates superior performance, achieving higher accuracy than the re-finetuned UmURL under the same experimental conditions.
In addition, other experimental results also show that our method achieves significant improvements over other methods, demonstrating exceptional performance in scenarios with limited labeled data.

\begin{table}[thb]
\caption{A comparison with the mainstream methods in semi-supervised learning. }
\centering 
\scalebox{0.9}{
\begin{tabular}{@{}l*{6}c @{}}
\toprule
\multirow{2}{*}{\textbf{Method}}   &
\multirow{2}{*}{\textbf{Modality}}   &
\multicolumn{2}{c}{\textbf{x-sub}} & \multicolumn{2}{c}{\textbf{x-view}} \\

\cmidrule(r){3-4} \cmidrule(r){5-6} 
&& 1\%  & 5\%  & 1\%  & 5\%  \\
\hline
ISC \cite{thoker2021skeleton} & J  & 35.7 & 59.6 & 38.1 & 65.7  \\
MCC \cite{su2021self} & J  & - & 47.4 & - & 53.3 \\
Hi-TRS \cite{chen2022hierarchically} & J  & 39.1 & 63.3 & 42.9 & 68.3 \\
GL-Transformer \cite{10.1007/978-3-031-19772-7_13}  & J  & - & 64.5 & - & 68.5 \\
HiCo \cite{dong2023hierarchical} & J  & 54.4 & - & 54.8 & - \\
CPM \cite{10.1007/978-3-031-19772-7_3} & J  & \underline{56.7} & - & 57.5 & - \\
CMD \cite{mao2022cmd} & J  & 50.6 & \underline{71.0} & 53.0 & \underline{75.3} \\
UmURL \cite{10.1145/3581783.3612449}&J&52.7&70.1&\underline{55.2}&73.6\\
Ours&J&\textbf{57.2}&\textbf{74.0}&\textbf{59.1}&\textbf{76.9}\\
\hline
AimCLR \cite{guo2022contrastive} & J+M+B  & 54.8 & - & 54.3 & - \\
CMD  \cite{mao2022cmd}& J+M+B  & \underline{55.6} & \underline{74.3} & 55.5 & \underline{77.2} \\
UmURL \cite{10.1145/3581783.3612449}&J+M+B&54.8&72.0&\underline{56.4}&75.6\\
Ours&J+M+B&\textbf{57.9}&\textbf{75.1}&\textbf{58.6}&\textbf{78.6}\\
\bottomrule
\end{tabular}
}
\label{tab:semisupervised}
\end{table}


\noindent\textbf{Transfer Learning for Skeleton-Based Action Recognition}
To validate whether the model can learn generalized features through contrastive learning and transfer the feature representation capability from one domain to another, we pretrain on one dataset and then fine-tune on another dataset, observing the accuracy of action classification.
Specifically, we conducted experiments where models pretrained on the NTU-60 and NTU-120 datasets were fine-tuned on the PKU-MMD II dataset, following the cross-subject (x-sub) protocol.
The experimental results in Table \ref{tab:transfer} demonstrate that our method also exhibits superior performance in transfer learning compared to previous works, highlighting the excellent transferability of the model.

\begin{table}[tb]
	\caption{A comparison with the mainstream methods in transfer learning.}
	\renewcommand{\arraystretch}{1.2}
	\centering 
	\scalebox{0.8}{
		\begin{tabular}{@{}l*{4}c @{}}
			\toprule
			\multirow{2}{*}{\textbf{Method}}   & 
			\multirow{2}{*}{\textbf{Modality}}   & 
			\multicolumn{2}{c}{\textbf{Transfer to PKU-MMD II}} &\\
			\cmidrule{3-4} 
			&& \multicolumn{1}{c}{\textbf{NTU-60}} &
			\multicolumn{1}{c}{\textbf{NTU-120}} &\\
			\hline
			LongT GAN  \cite{zheng2018unsupervised} & J & 44.8 & - \\
			MS$^2$L \cite{10.1145/3394171.3413548} & J & 45.8 & - \\
			ISC \cite{thoker2021skeleton} & J &  45.9 & - \\
			HiCo \cite{dong2023hierarchical} & J & 56.3 & 55.4 \\
			CMD \cite{mao2022cmd} & J & 56.0 & 57.0 \\
			UmURL \cite{10.1145/3581783.3612449} & J & \underline{58.2} & \underline{57.6} \\
			Ours & J & \textbf{59.2} & \textbf{59.0} \\
			\hline
			ActCLR \cite{lin2023actionlet}&J+M+B&55.9&-\\
			UmURL \cite{10.1145/3581783.3612449}& J+M+B & 59.7 & 58.5 \\ 
			Ours & J+M+B & \textbf{62.0} & \textbf{61.6} \\
			\bottomrule
		\end{tabular}
	}
	\label{tab:transfer}
\end{table}


\subsection{Ablation Studies and Visualizations}

\noindent\textbf{Effectiveness of Decomposition and Composition}
Table \ref{tab:ablation2} shows the impact of different strategies for decomposition and composition. When decomposition is spatial-temporal decoupled, the performance of the model improves. Adding feature composition also enhances the results. However, relying solely on composition is insufficient as composition focuses more on modality-invariant information and cannot learn enough modality-specific information.
Additionally, learning with spatial, temporal, and global features simultaneously does not further improve the results. 

\begin{table}[tb]
	\caption{Ablation studies on different strategies for decomposition and composition. `G', `T', and `S' represent global, temporal, and spatial representations respectively.}
	\centering
	\begin{tabular}{cccccc}
		\toprule
		\multicolumn{2}{c}{\textbf{Decomposition}} & \multicolumn{2}{c}{\textbf{Composition}} & \multirow{2}{*}{\textbf{x-sub}} & \multirow{2}{*}{\textbf{x-view}} \\ 
		\textbf{G} & \textbf{S + T} & \textbf{G} & \textbf{S + T} & ~ & ~ \\ 
		\hline
		\checkmark& - &\checkmark&-&84.8&91.5\\
		-&\checkmark&-&-&85.3&91.7\\
		-&-&-&\checkmark&82.1&88.1\\
		-&\checkmark&-&\checkmark&85.8&91.8\\
		\checkmark&\checkmark&\checkmark&\checkmark&85.5&91.9\\
		\toprule
	\end{tabular}
	\label{tab:ablation2}
\end{table}

\begin{figure}[tb]
	\centering
	\begin{subfigure}[b]{0.23\textwidth}
		\centering
		\includegraphics[width=\textwidth]{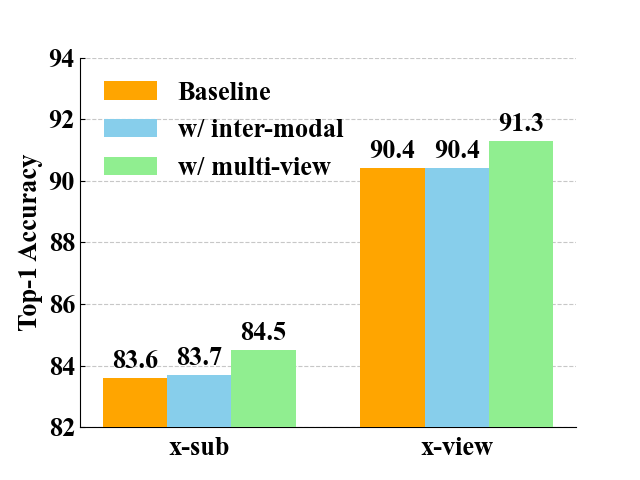} 
		\caption{}
		\label{fig:abl}
	\end{subfigure}
	\hfill
	\begin{subfigure}[b]{0.23\textwidth}
		\centering
		\includegraphics[width=\textwidth]{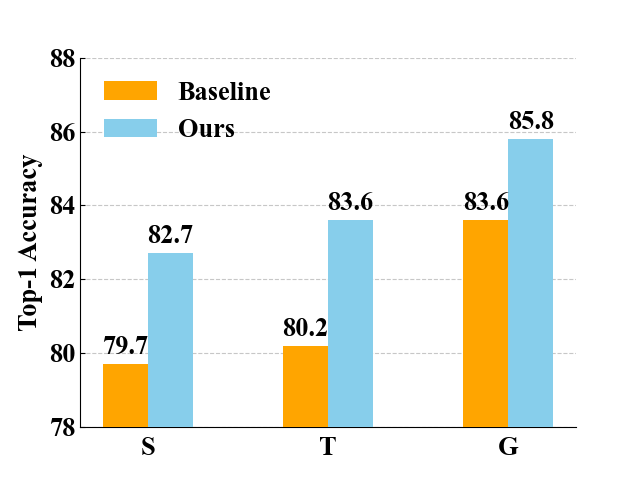} 
		\caption{}
		\label{fig:stg}
	\end{subfigure}
	\caption{(a) Impact of the inter-modal consistency loss and multi-view training. (b) Comparison of action recognition results on NTU-60 of x-sub protocol with different features.}
	\label{fig:abl_stg}
\end{figure}

\begin{figure}[tb]
	\centering
	\begin{subfigure}[b]{0.23\textwidth}
		\centering
		\includegraphics[width=\textwidth]{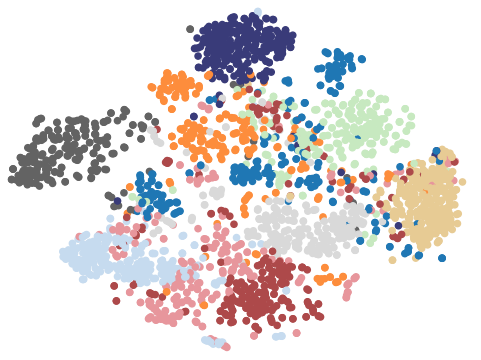} 
		\caption{Baseline}
	\end{subfigure}
	\hfill
	\begin{subfigure}[b]{0.23\textwidth}
		\centering
		\includegraphics[width=\textwidth]{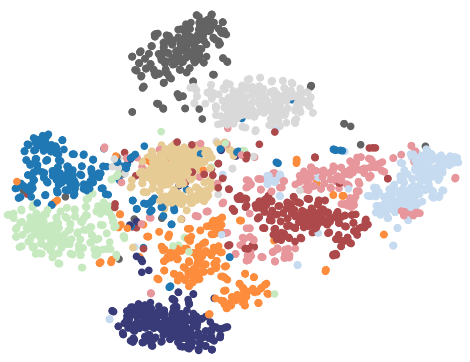} 
		\caption{Ours}
	\end{subfigure}
	\caption{Comparisons of learned features from 10 randomly selected different categories between the baseline and ours.}
	\label{fig:vis_feat}
\end{figure}

\noindent\textbf{Impact of Inter-Modal Consistency}
For unified multimodal pretraining, we do not utilize the inter-modal consistency loss, which is commonly employed in contrastive self-supervised learning. Fig. \ref{fig:abl_stg}(a) shows the ablated results of the inter-modal consistency loss. 
We find that adding inter-modal consistency loss to the baseline has little effect on the results, while this loss is incorporated into our approach, the results even decrease.
It can be explained that Unimodal Feature Decomposition in the baseline implicitly incorporates inter-modal alignment, so explicitly adding this loss is unnecessary when introducing feature decomposition and composition. 

\noindent\textbf{Impact of Multi-View Training} 
As is shown in Fig. \ref{fig:abl_stg}(a), when viewpoint-invariant training is introduced, the performance improves by approximately 1\% compared to the baseline.
Viewpoint Invariant Training successfully enhances the feature representation and improves the model's adaptability to varying viewpoints, resulting in higher accuracy and robustness in skeleton action recognition.

\noindent\textbf{Impact of and Global, Spatial and Temporal Features}
Fig. \ref{fig:abl_stg}(b) illustrates the significant differences in action recognition accuracy between our method and the baseline with feature types. Our method consistently achieves higher accuracy than the baseline for spatial, temporal, and global features. 
The results clearly prove the substantial impact the decoupled decomposition and composition have on both spatial and temporal features. At the same time, as the quality of temporal and spatial features improves, the overall performance of global features also becomes better.

\begin{table*}[thb]
\centering 
\caption{ A detailed comparison between our method and the baseline under different modality selection strategies during the inference stage.
}
\begin{tabular}{lccccccc}
        \toprule
        Method &  Modality & \multicolumn{2}{c}{NTU-60} & \multicolumn{2}{c}{NTU-120} & \multicolumn{1}{c}{PKU-MMD II} \\
        \cmidrule(lr){3-4} \cmidrule(lr){5-6} \cmidrule(lr){7-7}
           &  & x-sub & x-view & x-sub & x-setup & x-sub \\
        \midrule
        Baseline&J&82.0&89.5&73.7&74.5&52.6\\
        Baseline&M&78.9&85.6&69.2&69.5&43.6\\
        Baseline&B&81.5&88.7&73.7&74.6&52.1\\
        Baseline&J+M&82.6&89.5&71.2&76.0&54.5\\
        Baseline&J+B&82.0&89.6&75.1&75.2&53.2\\
        Baseline&M+B&82.1&88.8&70.5&74.6&52.7\\
        Baseline&J+M+B&83.6&90.4&73.2&76.2&54.7\\
        \midrule
        Ours&J&84.1&90.8&75.2&76.7&52.7\\
        Ours&M&79.8&86.7&69.6&71.7&44.4\\
        Ours&B&84.3&90.5&75.9&77.6&54.4\\
        Ours&J+M&84.5&90.7&75.9&77.3&53.0\\
        Ours&J+B&84.3&91.3&75.9&77.7&55.6\\
        Ours&M+B&84.1&90.0&75.2&76.6&50.3\\
        Ours&J+M+B&85.8&91.8&77.5&78.8&54.7\\
        \bottomrule
    \end{tabular} 
\label{tab:linear_detailed}
\end{table*}

\noindent\textbf{Results of Different Modalities}
In Table \ref{tab:linear_detailed}, we provide a detailed comparison of the results of different modalities across various datasets during the prediction stage using the linear protocol. 
This serves as a comprehensive comparison and thorough demonstration of the effectiveness of the proposed method.
Our approach outperforms the baseline \cite{10.1145/3581783.3612449} by a significant margin. 
The accuracy of modality J even reaches the level of the baseline that used the combination of joint, motion, and bone modalities, demonstrating the great potential of our method.

In terms of modality selection, the overall performance of the model improves as the number of modalities used increases. This is because the addition of modalities provides extra information as well as complementary information between modalities. 
When using single modality features, the results of modality J outperform others. Similarly, when using multimodal features, the combinations that include the joint modality also perform better than other combinations. Therefore, incorporating the joint modality is reasonable in action recognition.
Notably, in the more complex PKU-MMD II dataset~\cite{liu2020benchmark}, the performance of motion representation in skeletal sequences is significantly inferior to that of other modalities and may have negative impacts on features involving motion modalities.
This indicates that multi-modal skeletal learning will still require improvement when dealing with complex datasets in the future.

\noindent\textbf{Visualizations} 
We also use t-SNE to visualize the learned features in Fig.~\ref{fig:vis_feat}. Compared to the unified multimodal pretraining baseline, our method can more effectively cluster features of the same action together while discriminating features across different actions.

\section{Conclusion}
In this paper, we present an approach for skeleton-based action understanding based on unified multimodal pretraining. The proposed Decomposition and Composition training framework incorporates Decoupled Spatial-temporal Unimodal Feature Decomposition, Multimodal Feature Composition, and Viewpoint-Invariant Training to enhance the performance of model. 
Extensive experiments on the NTU RGB+D 60, NTU RGB+D 120, and PKU-MMD II datasets show that our method achieves satisfying performance with low computational cost. 
These results demonstrate that the Decomposition and Composition framework achieves an excellent balance between accuracy and efficiency, even surpassing state-of-the-art methods that rely on the rule-of-thumb late fusion.

\noindent{\textbf{Limitation Discussion}} 
Similar to most existing works, this study concentrates on learning action representations from laboratory datasets. However, self-supervised learning utilizing more realistic, out-of-laboratory skeleton data remains an area to be explored in the future. While datasets such as NTU RGB+D offer valuable benchmarks, they still fall short in terms of the diversity found in noisy and complex data typically encountered in real-world scenarios.

\section*{Acknowledgments}
This work was supported by National Science Foundation of China (62172090, 62302093), Jiangsu Province Natural Science Fund (BK20230833), Start-up Research Fund of Southeast University under Grant RF1028623097, and Big Data Computing Center of Southeast University.



\bibliography{main}



\end{document}